\pdfoutput=1

\documentclass[11pt]{article}

\usepackage{acl}

\usepackage{times}
\usepackage{latexsym}

\usepackage[T1]{fontenc}

\usepackage[utf8]{inputenc}

\usepackage{microtype}

\usepackage{inconsolata}

\usepackage{graphicx}

\usepackage{subcaption}
\usepackage{multicol}
\usepackage{makecell}
\usepackage{tikz}
\usepackage{arydshln}
\usepackage{array} 
\usepackage{multirow}
\usepackage{listings}
\usepackage{svg}

\usepackage{xcolor}
\usepackage{longtable}
\usepackage{algorithm}
\usepackage{algpseudocode}
\usepackage{amsmath}
\usepackage{bbm}
\usepackage{float}
\usepackage{amssymb}
\usepackage{cleveref}
\usepackage{booktabs}
\newcommand{\Hquad}{\hspace{0.5em}} 
\usepackage{mathtools}
\usepackage{placeins}

\usepackage{enumitem}
\setlist[itemize]{align=parleft,left=0pt..1em,itemsep=0pt}
\definecolor{mycolor}{HTML}{fff6c2}

\makeatletter
\algrenewcommand\ALG@beginalgorithmic{\footnotesize}
\makeatother

\usetikzlibrary{calc}

\usepackage{pgfplots}
\pgfplotsset{compat=1.12}

\newcommand{\palmlarge}{\textsc{PaLM-2}}
\newcommand{\structsum}{\textsc{StructSum}}
\newcommand{\autoqa}{\textsc{Auto-QA}}

\title{\structsum{} Generation for Faster Text Comprehension}

\author{Parag Jain$^1$\thanks{~~Work done while interning at Google DeepMind.}\hspace{0.2cm}, Andreea Marzoca$^2$, Francesco Piccinno$^2$ \\
 School of Informatics, University of Edinburgh$^1$\\
Google DeepMind$^2$\\
\texttt{parag.jain@ed.ac.uk},  \texttt{\{andreeam,piccinno\}@google.com}}

\begin{document}
\maketitle
\begin{abstract}
We consider the task of generating structured representations of text using large language models (LLMs). We focus on tables and mind maps as representative modalities. Tables are more organized way of representing data, while mind maps provide a visually dynamic and flexible approach, particularly suitable for sparse content. Despite the effectiveness of LLMs on different tasks, we show that current models struggle with generating structured outputs. In response, we present effective prompting strategies for both of these tasks.  We introduce a taxonomy of problems around factuality, global and local structure, common to both modalities and propose a set of critiques to tackle these issues resulting in an absolute improvement in accuracy of $+37$pp ($79\%$) for mind maps and $+15$pp ($78\%$) for tables. To evaluate semantic coverage of generated structured representations we propose \textsc{Auto-QA}, and we verify the adequacy of \textsc{Auto-QA} using SQuAD dataset. We further evaluate the usefulness of structured representations via a text comprehension user study. The results show a significant reduction in comprehension time  compared to text when using table ($42.9\%$) and mind map ($31.9\%$), without loss in accuracy.

\end{abstract}
\section{Introduction}
\begin{figure}[t]
    \includegraphics[width=\linewidth]{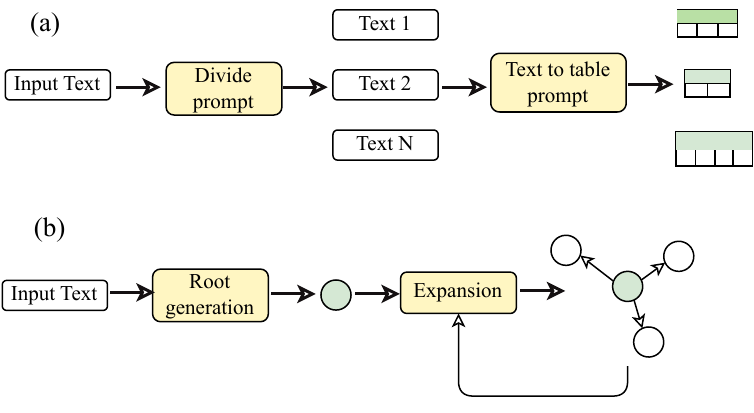}
    \caption{Overview of (a) tables and (b) mind map generation prompts. The prompting steps are \colorbox{mycolor}{colored}. Figure (a) illustrates the divide-and-generate prompt. The input passage is initially segmented into sub-passages, followed by the generation of multiple tables. Figure (b) demonstrates the generation process for mind maps. After the main concept has been generated, an iterative expansion phase ensues, during which the mind map is expanded until termination.}
    \label{fig:overview}
\end{figure}

The overwhelming amount of information available online poses a significant challenge for users seeking to quickly grasp and process relevant information. Current large language models (LLMs), such as \palmlarge~\citep{PaLM2}, Gemini~\citep{gemini23gemini} and ChatGPT~\citep{openai2022chatgpt}, while capable of providing text-based responses to user queries, often fail to adequately structure and organize this information in a way that facilitates comprehension~\citep{tang2023strucbench}. This can lead to information processing bottlenecks that hinder users' ability to efficiently extract meaningful insights from text.

\begin{figure*}[t]
    \centering
    \includegraphics[width=1\textwidth]{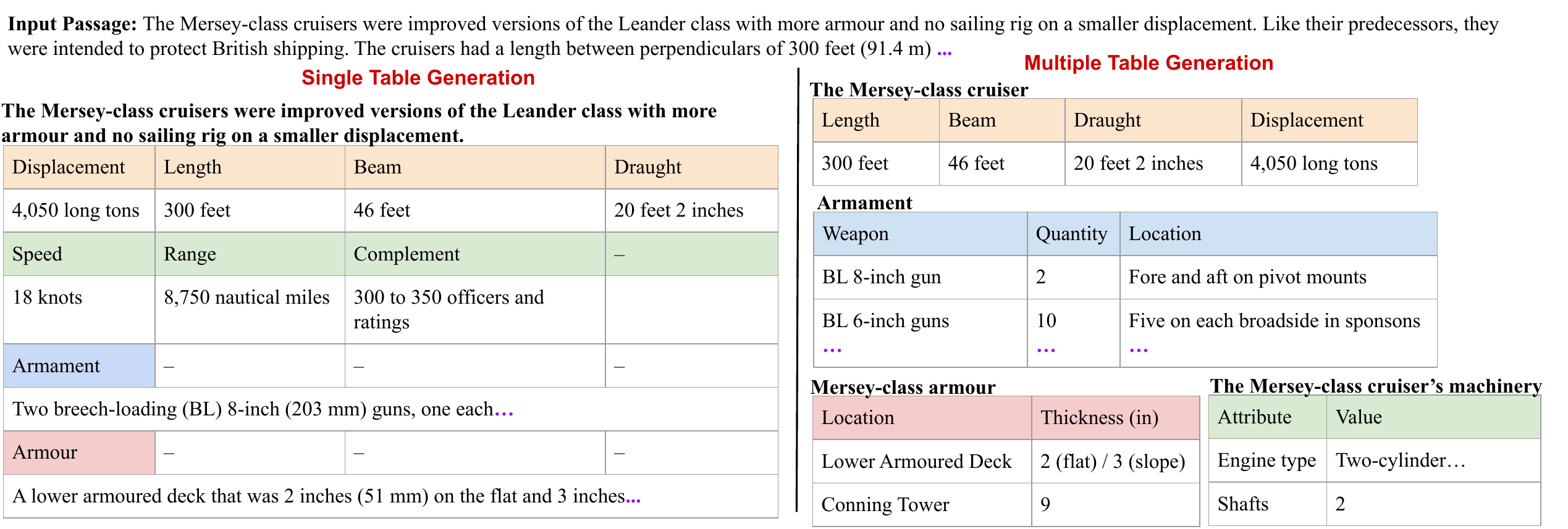}
\caption{Example table generation for the text at top, comparing single table (left) vs multiple table generation (right). Some parts in the table and text were truncated (\textcolor{purple}{...}) for readability. The full example is reported in Figure~\ref{fig:full_table_example}.}
    \label{fig:table-example}
\end{figure*}

\noindent To address this issue, we introduce the notion of structured summaries, or \structsum{} in short. \structsum{}s are derived by hierarchically organizing information and inducing semantic connections from an input text passage.  Without loss of generality, we focus on tables \citep{wu-etal-2022-text-table, li-etal-2023-sequence-sequence} and mind maps~\citep{buzan, huang-etal-2021-efficient} as possible \structsum{} instantiations:

\begin{itemize}
    \item \textbf{Tables} are well-studied in the NLP literature. However the vast majority of the work focused on simpler tasks where tables are inputs -- such as QA~\citep{herzig-etal-2020-tapas}, semantic parsing~\citep{bogin-etal-2019-global}, NLG~\cite{andrejczuk-etal-2022-table, 10.1162/tacl_a_00381, 10.1162/coli_a_00363}, etc. -- rather than outputs. Indeed, faithfully transforming an arbitrary text passage into a table is a difficult task as the model must deal with different challenges, such as reasoning at multiple levels, dealing with missing information, and visually consistent formatting. Motivated by the limitations above, we propose to generate multiple tables instead. We argue that this is a simpler task for an LLM, as shown in \Cref{fig:table-example}, which compares single-table and multi-table generation side by side. We therefore propose a divide-and-generate prompting approach (see Figure~\ref{fig:overview}) that first divides the input text into multiple text passages, each representing a sub-topic, %
    followed by an LLM prompt to generate a table-caption pair for each smaller passage. This decomposition allows the model to generate smaller, focused and more informative tables, especially for complex text passages with multiple sub-topics.
    
    \item \textbf{Mind maps}~\citep{hu-etal-2021-efficient,ijcai2019p729} are less studied in the literature, but are  helpful for comprehension and learning~\citep{buzan, Dhindsa2011}. Mind maps are complementary to tables in their structure, allowing for more flexibility and dynamism than tables, as they are inherently schema-less. However, generating mind maps with LLMs presents several challenges: %
    (i) the model first need to select a central concept, that is the fulcrum of all the successive extractions, as mind maps revolve around a central root node;
    (ii) being a schema-less abstraction, each connecting branch has its own independent sub-topic, making it difficult to automatically add branches all at once;
    (iii) to ensure readability and well-structuredness each leaf node should terminate the path in a way that concludes the idea or sub-topic;
    (iv) depending on the information density, some paths may be shorter than others. Therefore, the model should decide whether or not a branch is worth expanding.
    Following the structure of these observations, we propose an iterative prompting technique for mind map generation. As show in Figure~\ref{fig:overview}, we initialize the mind map by generating the root concept. At each iteration, we decide either to expand the current mind map further or stop the process. During the expansion step, we prompt the model to add branches to the current leaf nodes. We represent the mind map as a JSON object, as it is easy to parse and verify.

\end{itemize}

\noindent Through extensive experimentation with \palmlarge~\citep{PaLM2}, we show that LLMs are not always effective at  generating \structsum{}s that  are factual and structurally correct. To overcome these issues we propose a pipeline for structured data  generation. Our pipeline consists of structure-specific prompts followed by critics to assess output quality along three different dimensions, that are common both to tables and mind maps: (i) \emph{Factuality}, (ii) \emph{Local Structure} and (iii) \emph{Global Structure}. We found  that  our  proposed  critics  improved the overall quality of the generated output by +$37$pp for mind maps and +$15$pp for tables.

To ensure the usefulness of \structsum{} for text-comprehension tasks, we propose Auto-QA as a measure of  output  coverage. We  automatically generate QA pairs from input text and use structured outputs to answer these questions. Furthermore, we verify the appropriateness of using Auto-QA by comparing Auto-QA with human generated QA pairs on SQuAD~\citep{rajpurkar-etal-2016-squad} development set.

Finally, starting from the initial hypothesis that \structsum{}s can enhance the effectiveness of information-seeking scenarios, we conducted a user study to evaluate their impact on users' ability to process information, using a text comprehension user study. Results demonstrate how \structsum{}s improve information seeking, specifically on timed text comprehension metrics. We found that by using the structured representation, users can answer questions $42.9\%$ faster for tables and $31.9\%$ for mind maps.

\section{Related Work}
\paragraph{\textbf{Structured Output.}}
Generating structured output from text has been explored in the context of information extraction~\citep{li-etal-2023-sequence-sequence,pietruszka2022stable}. Most of the work focus on text-to-table~\citep{wu-etal-2022-text-table} generation using the model trained on domain specific dataset.~\citet{ni2023unified} use LLM for information extraction by generating key-value pairs. ~\citet{tang2023strucbench} evaluate different models on table generation from text by prompting where table structure is provided as format instructions. Mind map generation has been explored in the form of relation graph structure~\citep{hu-etal-2021-efficient, ijcai2019p729} to summarize new articles~\citep{cheng-lapata-2016-neural, hermann2015teaching}. In contrast, we focus on a generation pipeline applicable for multiple structured outputs types by prompting LLM given a text input. We keep the output structure flexible and domain independent by not instructing the model with specific format.

\paragraph{\textbf{Prompting.}}
Our prompting strategy is rooted in task decomposition techniques. %
Least-to-most~\citep{zhou2023leasttomost}, in contrast with chain-of-thought~\citep{wei2022chain}, progresses from easiest to hardest questions eventually answering the complete question, while successive prompting~\citep{dua-etal-2022-successive} iteratively generate new questions based on previous answers. Unlike least-to-most, decomposed prompting~\citep{khot2023decomposed} doesn't restrict task decomposition from easiest to the hardest and iteratively generate next steps that can be executed by different systems. Most of the prior work is focused on reasoning for solving QA type problems, in contrast, we are interested in transforming text to structured formats. Our divide-and-generate prompting for multiple table generation (similar to least-to-most) uses an initial prompt to divide the input passage into different topical sub-passages that simplifies the table generation in next step. Different from these tasks our iterative prompting for mind maps requires reasoning over current structured output at each step.

\paragraph{\textbf{Factuality.}}
Attribution is used as a tool for assessing the reliability of LLMs and identifying potential sources of inaccuracy or fabrication in their generated outputs. Current work apply attribution on unstructured text generation settings, such as, question answering~\citep{https://doi.org/10.48550/arxiv.2212.08037} and text generation tasks~\citep{gao-etal-2023-rarr}. Diverging from that, our work require verifying the factuality of generated structured outputs.

\paragraph{\textbf{Evaluation.}}
Due to the cost of human evaluation, LLMs are used to critique the generated outputs~\citep{wang2023shepherd}. Recent instructions tuned models, such as, GPT-4~\citep{openai2023gpt4} and ChatGPT~\citep{openai2022chatgpt} are shown to be strong evaluators. To avoid using external APIs, ~\citet{kim2023prometheus,wang2023shepherd} fine-tune a smaller pretrained model to critic model responses. We are interested in evaluating the quality structured outputs using critics and self-correct based on the feedback. As a part of data generation pipeline, our focus is on filtering instances that are incomplete and are not factually grounded.

\section{Generating \structsum{}s}

We focus on tables and mind maps as a possible \structsum{} instantiations.

\subsection{Tables: Divide \& Generate prompting} 

Given an input text we would like to transform it into multiple tables. Although generating multiple tables from text may seem unnecessary, single-table generation lead to several issues, as shown in Figure~\ref{fig:table-example} (bottom left). The model often produces complex table structures, resulting in missing cell values or the exclusion of relevant information. Additionally, complex tables are difficult to verify for factual accuracy and can require additional mental effort from the user to understand.
\par To address these limitations, we propose a divide-and-generate approach that dynamically partitions the passage into smaller subtopic segments. While deterministic rule-based chunking methods (e.g., based on word or sentence count) can be employed, they often produce suboptimal results due to potential under-chunking, over-chunking, and the absence of division for certain instances. Therefore, the chunking must be adaptive and depend on the input text and its sub-topic distribution. We use a one-shot prompt for this step, as shown in Appendix~\ref{sec:prompts} (Figure ~\ref{fig:prompts/text_to_table_caption}). After the chunking, we prompt the model to generate a table along with its caption for each sub-passage obtained in the previous step.

\begin{algorithm}[th]
\small
\begin{algorithmic}[1]
\Require
\Statex $\text{input text passage: input }$
\Statex $\text{maximum number of steps: } \text{max\_steps}$
\State $\text{step} \gets \text{0}$
\State $\text{mindmap} \gets \Call{generate-root}{\text{input}}$
\While{$\text{step} < \text{max\_steps} $}
	\State $\text{step} \gets \text{step} + \text{1}$
	\If{\Call{continue-prompt}{\text{input}, \text{mindmap}}}
	    \State $\text{expansions} \leftarrow \Call{expand}{\text{input}, \text{mindmap}}$
	    \State $\text{mindmap} \leftarrow \Call{json-critic}{\text{expansions}}$
	\Else
	    \State \Return $\text{mindmap}$
	\EndIf
\EndWhile
\State \Return $\text{mindmap}$
\end{algorithmic}
\caption{Mind maps Iterative Prompting}
\label{alg:mm}
\end{algorithm}

\subsection{Mind maps: Iterative Prompting}
\begin{figure}[t]
    \centering
    \includegraphics[width=0.45\textwidth]{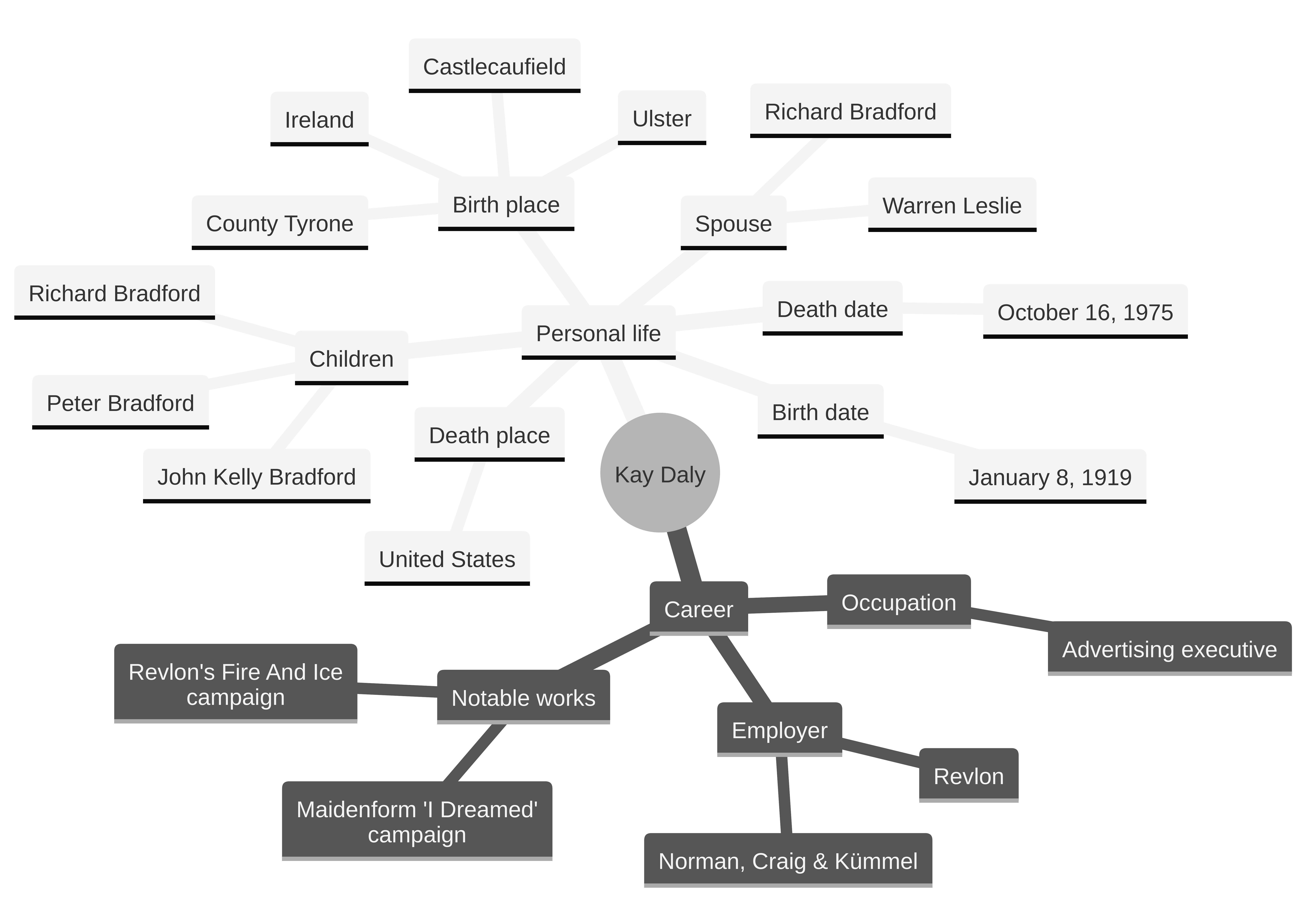}
    \caption{Example mind map output. The full example along with the input text is reported in Figure~\ref{fig:full_mindmap_example}.}
    \label{fig:mindmap-example}
\end{figure}

Contrary to tables, mind maps (see example in Figure~\ref{fig:mindmap-example}) are more flexible and present a different set of challenges. The first challenge is representation. We desire a representation that is (i) close to a familiar format, and (ii) is easily parsable and verifiable using current tools. JSON meets both of these requirements. The second challenge is that mind maps, unlike tables where each row can be produced linearly, necessitate attaching information in different locations depending on which branch is being expanded. This requires the model to think radially.

We propose an iterative prompting for mind maps generation. \Cref{alg:mm} shows the overall procedure. Details of each prompt is in \Cref{sec:prompts}. We start by generating the root concept that becomes the central node for the mind map. This separate step allows the model to independently reason about the theme of the passage. After generating the root, at each step we prompt the model to decide if current mind map can be expanded further. If the model decides to expand (line 5), we prompt the model using the current mind map to add more branches. Otherwise, the procedure terminates and we return the current mind map. At each expansion step we sample multiple mind maps. Utilizing the fact that JSON verification is cheaper we select the topmost JSON that is parsed correctly. In the rare case, when none of the samples are parsable we call a critic prompt to correct the top JSON (line 7).

\section{Data Generation Pipeline}
\label{sec:datapipeline}
We now present our \structsum{} data generation pipeline. Although each \structsum{} is seemingly different, we identify three dimensions that are common to both table and mind map modalities: (i) Factuality, (ii) Local Structure, and (iii) Global Structure. We use a set of critics, implemented via prompts, to ensure sufficient quality across each dimension. Through our initial experiments we find that tweaking each critic according to the structure is more helpful.

\subsection{Factuality Critic}
We use post-attribution~\citep{gao-etal-2023-rarr} to verify factuality, as we found that jointly generating and attributing~\citep{gao-etal-2023-enabling} results in (i) unnatural text output and (ii) in the model copying verbatim from the input text passage.

Critic cost is one aspect that requires consideration. For example, for tables, verifying each cell could be more robust, however, it increases the number of LLMs calls (listed in Table~\ref{tab:critic_cost}), from $\mathcal{O(\text{1})}$ to $\mathcal{O(\text{\#number of cells})}$.

For simplicity, we choose a single prompt per \structsum{}: for tables we ask the LLM to attribute each row, while for mind maps we ask to attribute each path from root to leaf. We convert the input text passage to a list of sentences and ask the model to cite, following the [x,y] format for attribution, the source sentence(s) where the information can be found or [NA] in case this is not possible.
The prompts are reported in \Cref{fig:prompts/critic_factuality}.

\subsection{Local Structure Critic}

For tables, a common issue arises from the model misplacing values in incorrect columns. For example, placing ``66 years'' in the \emph{Birth date} column or an address in the \emph{Company Name} column. To detect such errors, we leverage each column header as a category and verify whether all cell values within that column belong to the same category. For mind maps, we observed that a well-defined terminal node can often represent the entire path leading to it. We use this fact and prompt the model to verify whether the terminal node is a specific value, rather than a general concept.
The prompts are reported in \Cref{fig:prompts/critic_local_structure}.

\subsection{Global Structure Critic}
Global critic allows us to verify the overall structure of the output. This means understanding whether all the information contained in a \structsum{} makes sense globally.

\noindent For tables, we simply verify whether the table is well formatted: e.g. we verify equal number columns in the header and subsequent rows, therefore ignoring semantic content of the table and only focusing on form rather than the content. This is realized via simple heuristics implemented in Python (we do not prompt the model for these).

\noindent For mind maps, we used a stricter approach, to ensure that information were semantically valid on a global level. Specifically, we convert the mind map into a familiar format like table of contents (ToC), which we hypothesize is more likely to be seen during the pre-training phase of existing LLMs, and ask the model to check if the ToC is at right level of abstraction. The prompts are reported in \Cref{fig:prompts/critic_global_structure}.

\begin{table}[]
\centering
\begin{tabular}{lcc}
\toprule
\multirow{2}{*}{Critic} & \multicolumn{2}{c}{\# LLM calls} \\
                          & Tables    & Mindmaps  \\
\midrule
Factuality       & $\mathcal{O}(1)$      & $\mathcal{O}(1)$       \\
Local Structure  & $\mathcal{O}(\text{\#cols})$ & $\mathcal{O}(\text{\#paths})$ \\
Global Structure & NA        & $\mathcal{O}(1)$       \\
\bottomrule
\end{tabular}
\caption{Cost for each critic in terms of \#LLM calls as proxy. \#cols is number of columns in output table. \#paths is the number of paths in a mind map from root node to a terminal node.}
\label{tab:critic_cost}
\end{table}

\section{Semantic Coverage using \autoqa{}}
\label{sec:autoqa}

In this section we propose an automatic way to assess the quality and the general usefulness of \structsum{}s introducing \autoqa{} coverage as proxy metric.~\footnote{We do not present \autoqa{} as a substitute for human evaluations of quality. Instead, we propose AutoQA as a coverage metric that allows us to use synthetic question generation.}
This metric measures the semantic coverage or percentage of questions that are answerable when using a \structsum{} \Hquad $s$, instead of the full text passage $t$. Formally it is defined as:

$$COV(s) = \frac{1}{|\text{GenQA}(t)|} \sum_{i=1}^{|\text{GenQA}(t)|}{\mathbbm{1}_{E_{a_i}}\left[ Q(s, q_i)\right]}$$

\vspace{-\parskip} where $\text{GenQA}(x)$ is a function that generates $(q,a)$ pairs given the input text passage $t$, $Q(s, q_i)$ is a function that generates an answer given in input a \structsum{} $s$ and the question $q_i$, whereas the indicator function $\mathbbm{1}_{E_{a_i}}(x)$ asses the answer equivalence between $a_i$ and $x$. \Cref{fig:prompts/autoqa} in Appendix~\ref{sec:prompts}, show all the prompts associated with \textsc{AutoQA} module~\citep{10.1162/tacl_a_00397, fabbri-etal-2022-qafacteval}.

Independently of perceived quality, it is worth noting that this simple metric can be thought as an abstractiveness measure or compression quality for a given \structsum{} $s$.
A value of $1$ indicates no information loss at the expense of no compression/abstraction, whereas a value of $0$ indicates theoretically maximum compression at the expense of not providing any useful information. A target value is therefore application specific and must be adjusted accordingly~\footnote{It is possible to include coverage as a critic. But we opted not to do so, as the threshold for coverage depends on the specific use case. This also allowed us to analyze coverage independently, without being influenced by other factors.}.

\paragraph{QA pairs generation} \label{sec:qa_pair_gen} $\text{GenQA}(t)$ is implemented by prompting the LLM to generate a list of question-answer (QA) pairs conditioned on the input text $t$. To ensure that the quality of QA pairs is sufficient, after generation, we we apply a three-step procedure. First, we removed duplicate questions via string match. Second, we removed answers if none of the words appeared in the input text, thereby ensuring with reasonable certainty that the answer is grounded in the text without being overly stringent. Third, we performed a cyclic consistency check, where we prompted the model to answer the generated question based on input text.

\paragraph{Question answering} We use a simple prompt for function $Q(s, q_i)$. For tables, we convert the table representation to a markdown table format, whereas for mind maps we simply serialize the information as a JSON object.
\paragraph{Answer Equivalence} As the model might generate verbose answers, verifying whether two answers are the same is a problem of semantic similarity. Instead of using lexical matching, that is $\mathbbm{1}_{E_{a_i}}(x) := a_i = x$, we prompt the model to check if two answers are equivalent.

\section{Model}
For all the experiments, we use the Unicorn~\citep{PaLM-2-unicorn} variant of \palmlarge, a fine-tuned transformer-based model with UL2~\citep{tay2022ul2} like objectives. \palmlarge     ~ improves on PaLM~\citep{chowdhery2023palm} through optimized scaling, richer training data and instructing tuning~\citep{wei2021finetuned, chung2022scaling}.

\section{Dataset}
\label{sec:inputdata}
To test our pipeline on a diverse set of input passages, we selected Wikipedia text as the source. Specifically, we started with the English split of the \textsc{Wiki40b}~\citep{49029} dataset~\footnote{We used the version that is available via the Tensorflow datasets \url{https://www.tensorflow.org/datasets/catalog/wiki40b}.}. The dataset is cleaned up by page filtering to remove disambiguation pages, redirect pages, deleted pages, and non-entity pages. Input to our prompts are passages that are obtained by splitting the Wikipedia text using the \texttt{\_START\_PARAGRAPH\_} symbol that is already provided as part of the dataset.

\subsection{Filtering for Tables Generation} 
\label{sec:table_filtering}
Not all input paragraphs are well suited for table generation. As a proxy for selecting adequate passages, we used regex-based filters to only include passages with more that $20$ numeric values and removed passages with less than three sentences. In a real world setting, we would like a systematic way of deciding which modality is adequate for a given text. We leave this exploration as future work.
\section{Results}
In this section, we present the results of our experiments using \palmlarge.

\subsection{Quality impact of prompting style and automated critics}
\label{sec:quality_annotation}

We assessed the quality of generated structured data through manual human ratings. The study was conducted on $100$ instances for mind maps. For multi-table generation, we choose $100$ individual table-text pair for annotation~\footnote{We made sure that the input passages are the same for the different ablations within modalities. For multi-table generation, we choose $100$ text-table pairs generated using $52$ input passages.}. Input passages were obtained via data filtering strategy described in \Cref{sec:inputdata}.

\paragraph{Guidelines} Annotators were asked to rate each instance as ``Good`` or ``Bad'' by checking the overall quality of the output. For both modalities, annotators were asked to check for factuality as well as the structural quality of the output. To help the annotators measure the structural quality we asked the annotators to check ``table structure'', ``table header'', ``column header-value match'' for tables. For mind maps, they were asked to check ``incomplete branches'', ``not a good main concept'', ``too dense / too sparse'' and ``wrong edge connections''. We also encouraged the annotators to mark the instance as bad if they find any other issues.

\begin{table}
    \centering
    \begin{tabular}{lc}
    \toprule
    \multicolumn{2}{c}{\textbf{Tables}}\\
    \midrule
    Single Table   & 54 \\ %
    Multi Table    & 63 \\
    \bottomrule
    \end{tabular}
    \quad
    \begin{tabular}{lc}
    \toprule
    \multicolumn{2}{c}{\textbf{Mindmaps}}\\
    \midrule
    CoT         & 39 \\ %
    Iterative   & 42 \\
    \bottomrule
    \end{tabular}
    \caption{Table / mind map accuracy per prompt style. Outputting multiple tables provides higher quality for the table modality. For mindmaps, an iterative approach is to be preferred to a CoT approach. Full prompts are reported in the Appendix~\ref{sec:prompts}.}
    \label{tab:prompt_style_accuracy}
\end{table}
\begin{table}
    \centering
    \begin{tabular}{lcc}
    \toprule
    \textbf{Critic} & \textbf{Tables} & \textbf{Mind maps} \\
    \midrule
    Baseline$\dagger$    & 63 & 42 \\
    \Hquad $\xhookrightarrow{}$ Structure & 70 & 71 \\
    \Hquad\Hquad $\xhookrightarrow{}$ Factuality & 78 & 79 \\
    \bottomrule
    \end{tabular}
    \caption{Human annotation accuracy at different pipeline stages. The use of critics is a critical step to improve perceived quality. Local and Global Structure critic provides a significant lift for mind maps. The increase in performance for Factuality, is similar for both Tables and mind map.}
    \label{tab:critic_accuracy}
\end{table}
\paragraph{Prompt style} \Cref{tab:prompt_style_accuracy} show the results for both the modalities. For table generation task we find that annotators prefer multiple tables generation outputs compared to single table generation. This can be attributed to the fact that multi-table generation enables the model to generate more concise, focused and informative tables. For mind map, we compare chain-of-thought~\citep{wei2022chain} with our proposed iterative generation strategy described in \Cref{alg:mm}. We find that iterative generation were preferred over simpler prompt outputs.

\subsection{Do Critics Align with Human Ratings?}
\label{sec:critics}
Through our human annotations results in Section~\ref{sec:quality_annotation}, we find that many generated outputs are not of acceptable quality. To improve the quality of the generated data and to avoid costly human annotations, we propose to use a combination of critics as a measure of data quality. To verify the efficacy of our critics, we first filtered the generated dataset with our critics. Specifically, we performed a logical \texttt{AND} of individual critics and filtered the instances that do not pass the criterion. We then sampled 100 instances from filtered examples and conducted the same evaluation as in~\Cref{sec:quality_annotation}.

Results in \Cref{tab:critic_accuracy} show that using the proposed critics the overall quality is improved by a significant margin. We observe that data filtered using \textit{Structure} (Global and Local) and \textit{Factuality} critics improve the percentage of acceptable instances generated using the pipeline. We find that the quality of mind maps improve by absolute $+37pp$. Similarly, for tables quality improves by absolute $+15pp$. These results indicate that the critics were able to retain good examples and that the selection criterion is in agreement with human judgement.

\begin{figure}[t]
  \centering
    \begin{tikzpicture}[scale=0.76]
    \begin{axis}[xlabel={\textsc{Auto-QA} Coverage},ylabel={Percentage (\%)},tick pos=left, legend
        style={at={(0.6,.88)},anchor=west},xtick={10,20,30,40,50,60,70,80,90},ytick
        distance=10, ymax=100,
        nodes near coords=,]
    \addplot table[x index=0,y index=1,col sep=comma] {dataplot.dat};
    \addlegendentry{Tables}
    \addplot table[x index=0,y index=2,col sep=comma] {dataplot.dat};
    \addlegendentry{Mind map}

    \end{axis}
    \end{tikzpicture}
    \caption{\textsc{Auto-QA} based coverage. A point $\langle X,Y \rangle$ in each line show that $X$\% of data has at least $Y$\% of coverage measured using \textsc{Auto-QA}.}
    \label{fig:qa_acc}
\end{figure}

\subsection{Measuring Coverage via Auto-QA}
\label{sec:qa_results}
Results in \Cref{fig:qa_acc} show \textsc{Auto-QA} coverage for mind maps and tables. The curve shows for a particular coverage threshold what percentage of data meets that threshold. Overall, we observe that tables have better coverage compared to mind maps, meaning that they have an higher abstractiveness or information retention capacity.
Interestingly, even though both modalities are perceptually different, we notice that both of them follow similar trends.

\subsection{Is Auto-QA a reasonable metric?}

\begin{table}[]
\setlength\tabcolsep{0.1pt}
\begin{tabular*}{\linewidth}{@{\extracolsep{\fill}} lcc }
\toprule
& \multicolumn{2}{c}{QA Type} \\
\cline{2-3}
                            & Auto & Human \\
\hline
Mind map                                   & 55.6 & 61.4 \\
Multi-Table \small{(Divide-and-generate)}  & 66.8 & 69.3 \\
Single Table                               & 57.1  & 58.8 \\
Query Focused \small{(Single Table)}         & 81 & 85.5 \\
\bottomrule
\end{tabular*}
\caption{QA accuracy on different modalities as context, generated using SQuAD validation set. \textsc{Auto-QA} is automatic question-answer pair generation. Human QA are original SQuAD questions curated by humans.}
\label{tab:squad}
\end{table}

We investigate the feasibility of using \autoqa{} as a surrogate for manually written QA pairs. We aim to determine whether \autoqa{} can generate QA pairs of comparable quality to those written by humans, and leading to a similar evaluation of semantic coverage. To verify the same, we use randomly selected $1000$ \emph{<passage, question, answer>} triples from the SQuAD ~\cite{rajpurkar-etal-2016-squad} validation set (common for all the experiments). Using the text \emph{passage} as input we generate different \structsum{}s. Next, we generate a QA pair corresponding to each text \emph{passage}. This QA pairs acts as a substitute for human written QA pair for \autoqa{} study. The goal is to check whether, keeping the passage and output \structsum{} the same, there is a correlation in performance between human generated QA pairs and automatically generated QA pairs.

\Cref{tab:squad} shows the overall results. Second (Mind maps) and third (Multi-Table) row show the comparison between Human QA and Auto QA for our proposed divide-and-generate prompt for tables and iterative prompt for Mindmap generation. We can see that \autoqa{} has comparable results and is a reasonable substitute for human generated questions as a measure of semantic coverage. We further study the limitations of \autoqa{}, and the difference in Human vs Auto QA scores in Section~\ref{sec:limitations_autoqa}.
\begin{figure*}[th]
\begin{subfigure}{.5\textwidth}
  \centering
  \includegraphics[width=\linewidth]{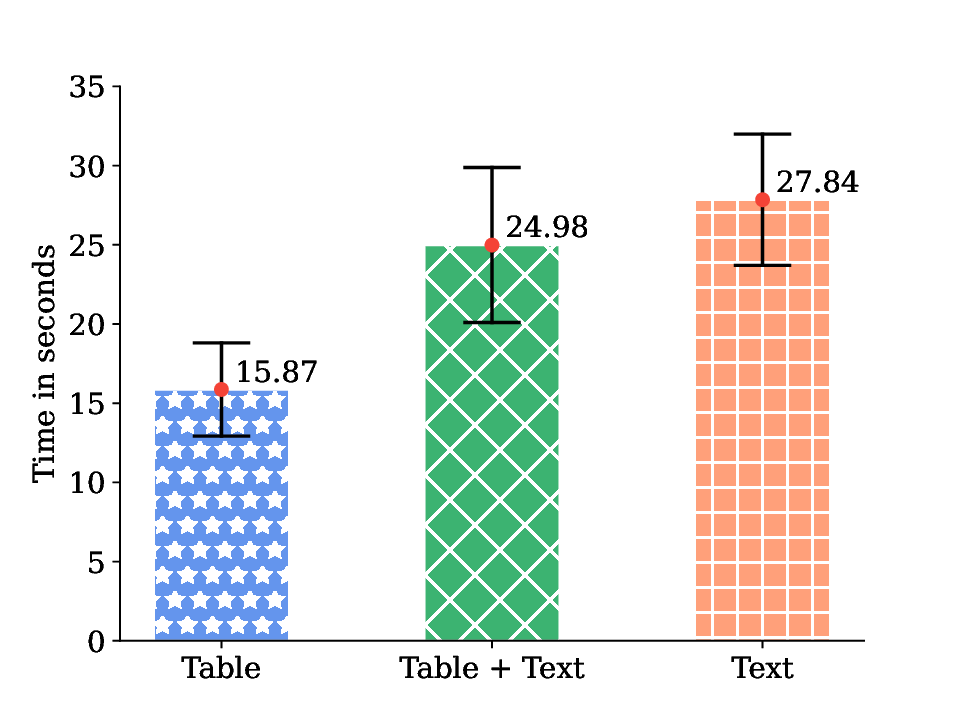}
  \caption{Table results}
  \label{fig:userstudyfig1}
\end{subfigure}%
\begin{subfigure}{.5\textwidth}
  \centering
  \includegraphics[width=\linewidth]{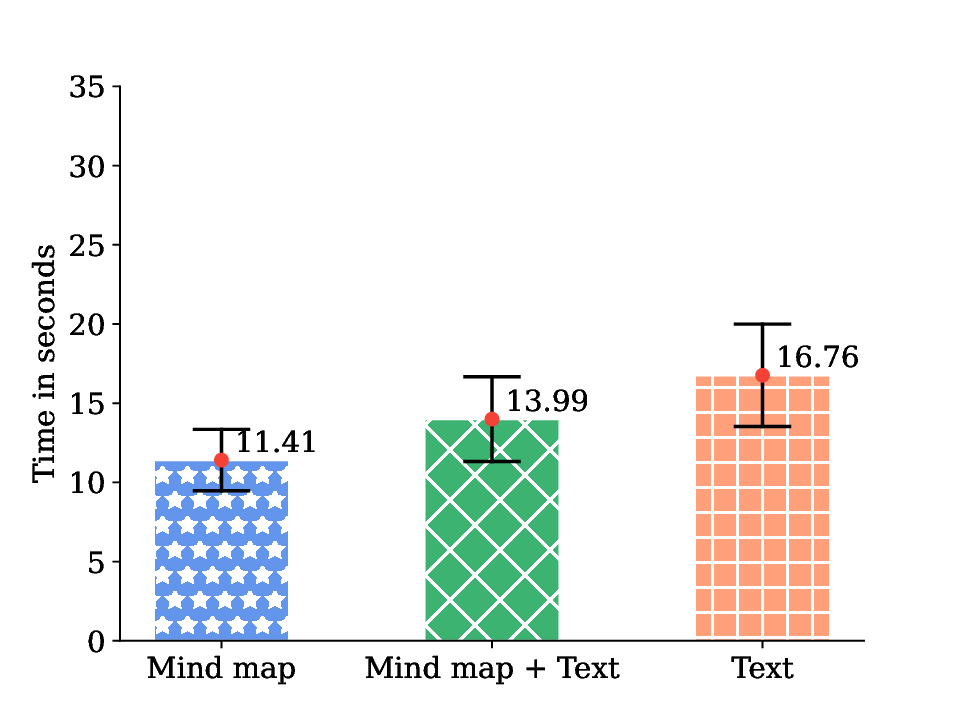}
  \caption{Mind map results}
  \label{fig:userstudyfig2}
\end{subfigure}%
\caption{Results for timed text comprehension based user study. Plots show 95\% confidence interval over time taken in seconds to answer question with different structure combinations as context. For both tables (left) and mind map (right), compared to text only, we observe significant reduction ($42.9\%$ and $31.9\%$ resp.) in average time taken by annotators to answer the question.}
\label{fig:userstudy}
\end{figure*}

\subsection{Multiple Tables vs Single Table} 

To check whether generating multiple tables is better at covering more information, we perform a comparison between the ability to answer questions by generating single or multiple tables. On comparing Multi-Table and Single Table row in \Cref{tab:squad}, we observe that for both \textsc{Auto-QA} and Human QA generating multiple table provides more coverage. So in addition to the benefits such as comparatively better verifiability and robust generation, multiple table generation are also better at covering more semantic information.

\subsection{Query Focused Generation}

In many cases user intent is known in advance, for example, a user query to search or LLM-based Assistant interface (e.g., ChatGPT, Gemini, etc.). We explore the possibility of generating structured data in the presence of a query. We perform a \textit{preliminary} analysis by adding the query in single table generation prompt. As we can see in last row in \Cref{tab:squad}, query focused generation improve the performance by more than $20$ for \textsc{Auto-QA} and $25$ points for Human-QA. Since this requires further investigation in terms of prompting and output quality analysis, we leave a comprehensive exploration of query-focused structured data generation as future work.
\begin{table*}[th!]
\centering
\small
\begin{tabular}{lcc}
\toprule
& \multicolumn{2}{c}{QA Type} \\
\cline{2-3}
                            & Auto & Human \\
\hline
Mind map                                   & 60.2 [57.0, 63.3]	& 61.3 [58.1, 64.4] \\
Multi-Table \small{(Divide-and-generate)}  & 72.3 [69.4, 75.2]	& 68.9 [66.0, 71.9] \\
Single Table                               & 61.8 [58.7, 64.9]	& 58.1 [54.9, 61.3] \\
Query Focused \small{(Single Table)}         & 87.7 [85.5, 89.8] & 85.5 [83.2, 87.8] \\
\bottomrule
\end{tabular}
\caption{Human and \autoqa{} accuracy with 95\% confidence interval on different modalities as context. Unlike the results presented in \Cref{tab:squad}, we have excluded SQuAD passages for which none of the questions generated by \autoqa{} passed the filter.} 
\label{tab:squad_filter_fix}
\end{table*}
\subsection{Are \structsum{}s useful?}
\label{sec:user_study}
We evaluate whether \structsum{} are useful abstractions for the users. For this we design a \textit{timed} text comprehension based user study. We assume that user is looking to answer a specific query, i.e. has a specific intent. We measure time taken to satisfy the user intent as a proxy of usefulness. Our evaluation team consists of 12 volunteers who are affiliated with our institution. Five of these volunteers are female, and seven are male. All volunteers are proficient in English, although not native speakers.
\par We create an intent in the form of a question along with different context combinations. For example, for a question $q$, we create $\langle q, s \rangle$, $\langle q, t \rangle$, $\langle q, s + t \rangle$ as possible combinations, where \emph{s} is a \structsum{} and $t$ is the input text passage. Each of these combinations are presented to different annotators while ensuring that no annotator see the same question twice. We then measure how long it takes to answer the question in each scenario.

\structsum{}s for the study were generated using our data generation pipeline and critic-based filtering, as discussed in Section~\ref{sec:datapipeline}. In total $600$ instances were annotated, equally divided into different context combinations for mind maps and table generation. Annotators consistently answered correctly across all context combinations (Appendix~\ref{app:user_study}), suggesting that the level of context did not significantly impact their accuracy.
\noindent\Cref{fig:userstudy} shows the overall results. The plots show 95\% confidence interval of time taken by the annotators when using different modality:
\begin{itemize}[noitemsep,nolistsep]
    \item \textbf{Tables.} \Cref{fig:userstudyfig1} shows that on average annotators with access to tables were able to answer almost $42.9\%$ time faster on average compared to annotators with only text. Furthermore, we observe that presenting both table and text is also useful to the annotators.
    \item \textbf{Mind maps.} \Cref{fig:userstudyfig2} shows the results for the study with mind map. A similar trend can be observed, with a reduction of approximately $31.9\%$ in average time between annotators with mind maps compared to annotators that only used text to answer the question.
\end{itemize}
\noindent We note that $\langle q, s + t \rangle$ performs worse than $\langle q, s \rangle$. We believe this is due to the fact that the annotators cross-checked the answer from both the modalities, leading to increase in time to answer the question.
\section{Data Generation statistics}
\label{app:sec:generation_stats}
\Cref{tab:table_stats} shows different statistics of data generated using our prompts. For tables generation we observe that our methods generate almost two ($\sim$ 1.9) tables per instance and the tables have 7.1 rows and 3.3 columns on average. Mind maps have an average of 11.8 nodes with a depth of 2.2. We show example mind map and table generation in Figure~\ref{fig:full_mindmap_example} and Figure~\ref{fig:full_table_example} respectively.

\begin{table}
\centering
\small
\begin{tabular}{ll}
\toprule
\multicolumn{2}{c}{\textbf{Tables}}\\
\midrule
Avg \#words per chunk     & 114.8  \\
Avg \#sentences per chunk & 3.9   \\
Avg \#words per input     & 240.6   \\
Avg \#sentences per input & 8.1   \\
Avg \#rows                & 7.1   \\
Avg \#cols                & 3.3   \\
Avg \#tables              & 1.9   \\
Max \#tables              & 11     \\
\toprule
\multicolumn{2}{c}{\textbf{Mind map}}\\
\midrule
Avg \#words & 194.6  \\
Avg \#sentences   & 7.9      \\
Avg \#nodes       & 11.8     \\
Avg depth         & 2.2     \\
\bottomrule
\end{tabular}
\caption{Table / mind map text input and output statistics. On average two ($\sim$ 1.9) tables (top) are generated per input text instance. Mind maps (bottom) contains 11.8 nodes on average.}
\label{tab:table_stats}
\end{table}

\section{Limitations of \autoqa{}}
\label{sec:limitations_autoqa}
We propose \autoqa{} as a coverage metric, in order to measure how much information from the original passage is retained in the \structsum{}. Note that the metric is not intended as a substitute for evaluations of overall quality. 
The primary benefit of \autoqa{} is the synthetic question generation component, which can be run at scale without costly human annotations. In Table~\ref{tab:squad}, we measured how  closely the synthetic \autoqa{} aligns with human generated question-answer pairs, revealing certain limitations of the metric. Notably the quality of generated questions is not always reasonable, which is mitigated using specialized filters, as discussed in Section~\ref{sec:qa_pair_gen}.  Although we generate several question-answer pairs, it is possible that none pass the filters, which we see for fewer than 10\% of input passages. Such instances adversely affect the \autoqa{} coverage score, contributing to the score discrepancies in Table~\ref{tab:squad}. When the analysis is restricted to those subsets where the \autoqa{} filtering is successful, the observed differences are diminished and fall within the bounds of experimental noise, as reported in \Cref{tab:squad_filter_fix}.

\section{Conclusion}
In this work we study the potential of structured representations like tables and mind maps to enhance information comprehension. Utilizing our divide-and-generate prompting and iterative expansion, we achieved significant improvements in output quality (+$37pp$ for mind maps, +$15pp$ for tables) using structure-specific prompts and critics. We proposed \autoqa{} based coverage metric that automatically generates QA pairs from the input text and uses \structsum{} outputs to answer them.

\section{Acknowledgment}
We would like to thank Massimo Nicosia and Yasemin Altun for their feedback and comments on the paper. We are grateful to all the volunteers who participated in our annotation experiments.
\section{Limitations}
We outline the limitations of our work to ensure transparency and inspire future research.
First, the structured output representations we experimented with are limited to tables and mind maps. However, to comprehensively evaluate the effectiveness of our critics and pipeline, it is desirable to also evaluate other input and output modalities, e.g. image and video, considering the recent advances in VLMs. 
Secondly, our work and experimental findings are limited to only English sources. We plan to also explore multilingual structured summaries in future work.
Third, we would to warn against the risk of blindly trusting models to generate structured summaries from an input accurately. Although we take extra care to increase the factuality of the outputs via the use of critics, and experimentally validate QA coverage, we believe that special care should be taken to verify outputs in accuracy-sensitive applications.
Finally, our \structsum{} generation is performed using a LLM with fixed prompts, however, prior work have shown a reasonable portability of prompts across similar models~\citep{zhou2023leasttomost, khot2023decomposed}.
\\ Despite these limitations, our work serves as an initial step in constructing reliable structured summarization evaluations, models and applications. We hope future research can greatly benefit from this starting point.

\bibliography{anthology,custom}
\appendix
\section{User Study}
\label{app:user_study}

\begin{figure*}
\centering
\begin{subfigure}{\textwidth}
    \texttt{Kathleen "Kay" Daly (January 8, 1919 – October 16, 1975) was an Irish-born American advertising executive and one of the four "celebrated Daly sisters". At Norman, Craig \& Kümmel she was the creative force behind the famous Maidenform "I Dreamed ..." campaign and Revlon's legendary 1952 Fire And Ice campaign, working with photographer Richard Avedon. She also was responsible for the line ''Every woman alive loves Chanel Number Five". She went on to join Revlon in 1961 as vice president and creative director. Kathleen Daly was born in Castlecaufield, County Tyrone, Ulster, Ireland, in 1919. Northern Ireland was created two years later with Tyrone one of its six counties. The family emigrated early in the 1920s. She grew up as one of four sisters, Maggie, Kay, Maureen, and American-born Sheila. They became known for their writing and work in journalism, fashion, and advertising, and were called "the celebrated Daly sisters" by Time magazine in 1966. Life magazine ran a feature story on them in 1949 and a follow-up in 1959. All four were at least once employed by the Chicago Tribune. When she moved to San Francisco after World War II, Kay Daly famously rented space on a billboard to advertise for an apartment. It not only netted her an apartment, but netted her nationwide fame and countless marriage proposals. She had a brief marriage to BMW executive and film producer Richard Bradford (part of the famous Bradford family of Plymouth Colony), who fathered her sons John (Kelly), Richard, and Peter. She then was married to journalist and executive Warren Leslie, who adopted and raised her sons, until her death on October 16, 1975, of pancreatic cancer. She was survived by husband Warren, sons Kelly, Peter, and Richard Bradford, and stepsons Warren and Michael Leslie.}
    \caption{Input text for mind map generation.}
    \end{subfigure}
   \begin{subfigure}{\textwidth}
    \includegraphics[width=\textwidth]{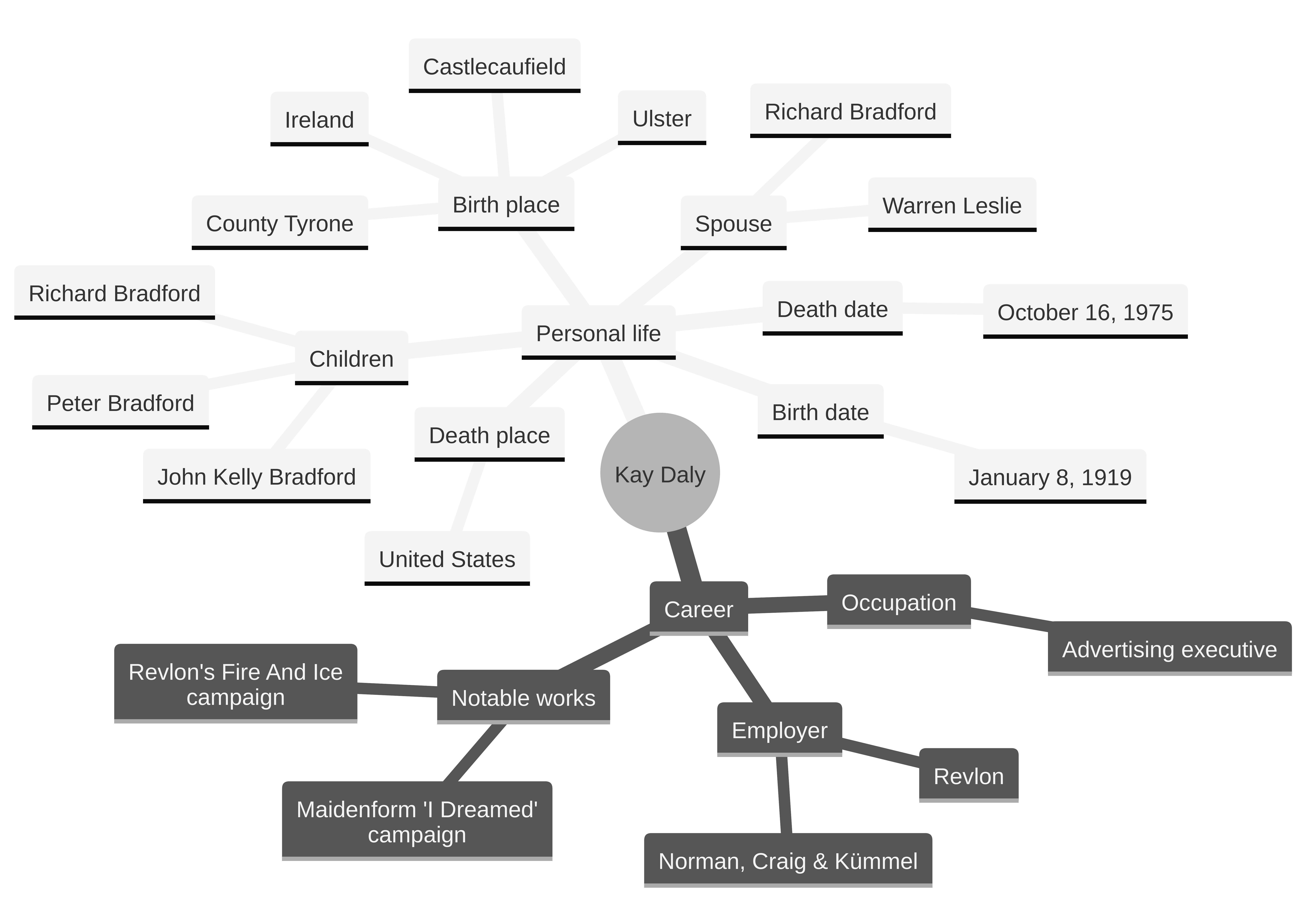}
    \caption{Mind map output.}
    \end{subfigure}
    \caption{Example mind map (below) generation for the input text (above). We use mermaid.js~(\url{https://mermaid.js.org/}) to visualize the output.}
    \label{fig:full_mindmap_example}
\end{figure*}

\begin{figure*}
    \centering
    \includegraphics[width=0.6\textwidth]{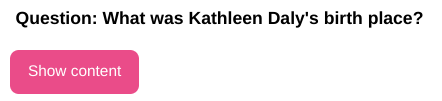}
    \caption{Example UI frame that is shown at the beginning of each annotation instance.}
    \label{fig:user_study_initq}
\end{figure*}

\begin{figure*}
    \centering
    \includegraphics[width=\textwidth]{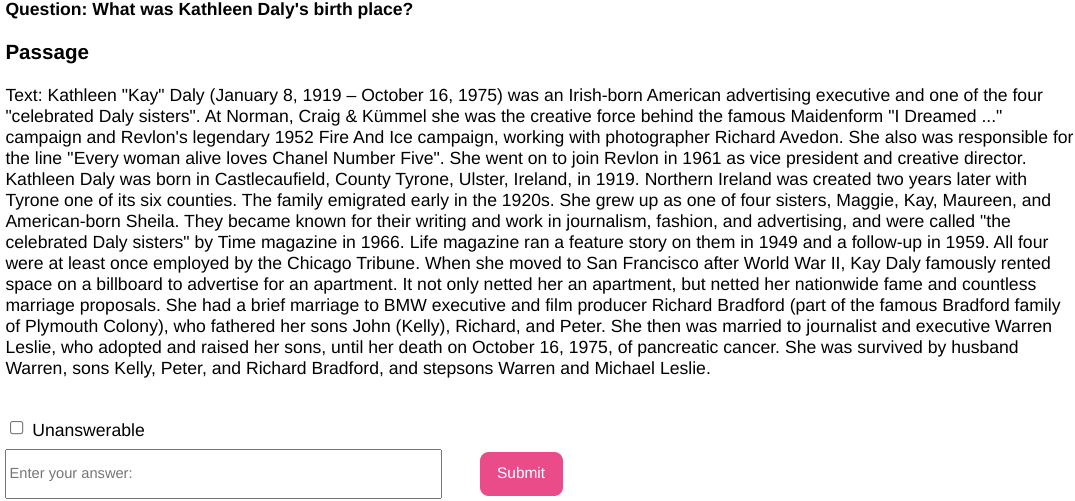}
    \caption{A followup frame shown after Figure~\ref{fig:user_study_initq} with text as context.}
    \label{fig:user_study_mmtext}
\end{figure*}

\begin{figure*}
    \centering
    \includegraphics[width=\textwidth]{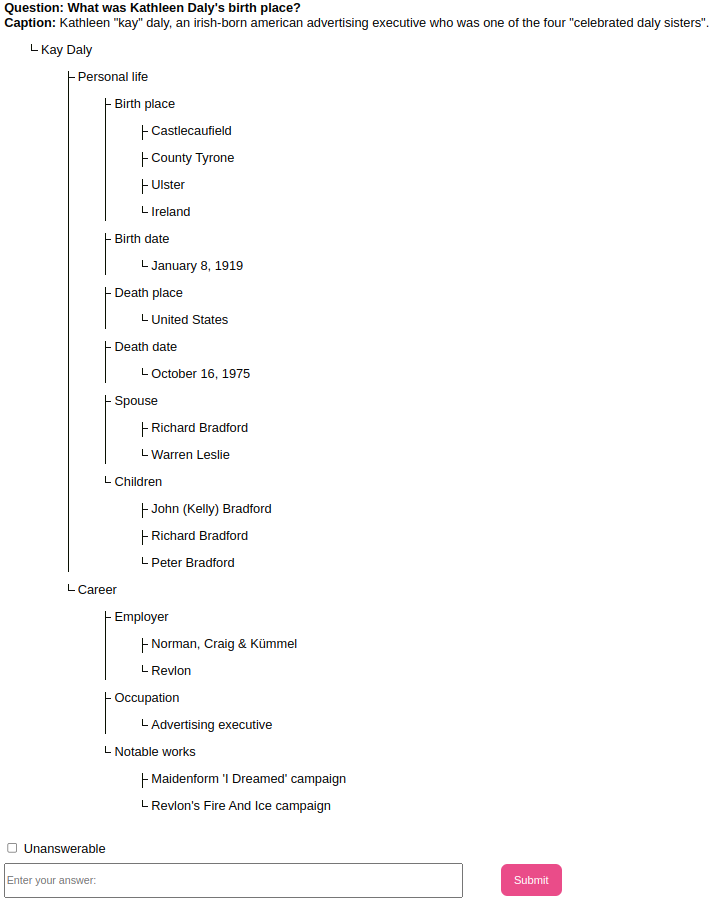}
    \caption{A followup frame shown after Figure~\ref{fig:user_study_initq} with structure (mind map) output as context.}
    \label{fig:user_study_mmonly}
\end{figure*}

\begin{figure*}
    \centering
    \includegraphics[width=\textwidth]{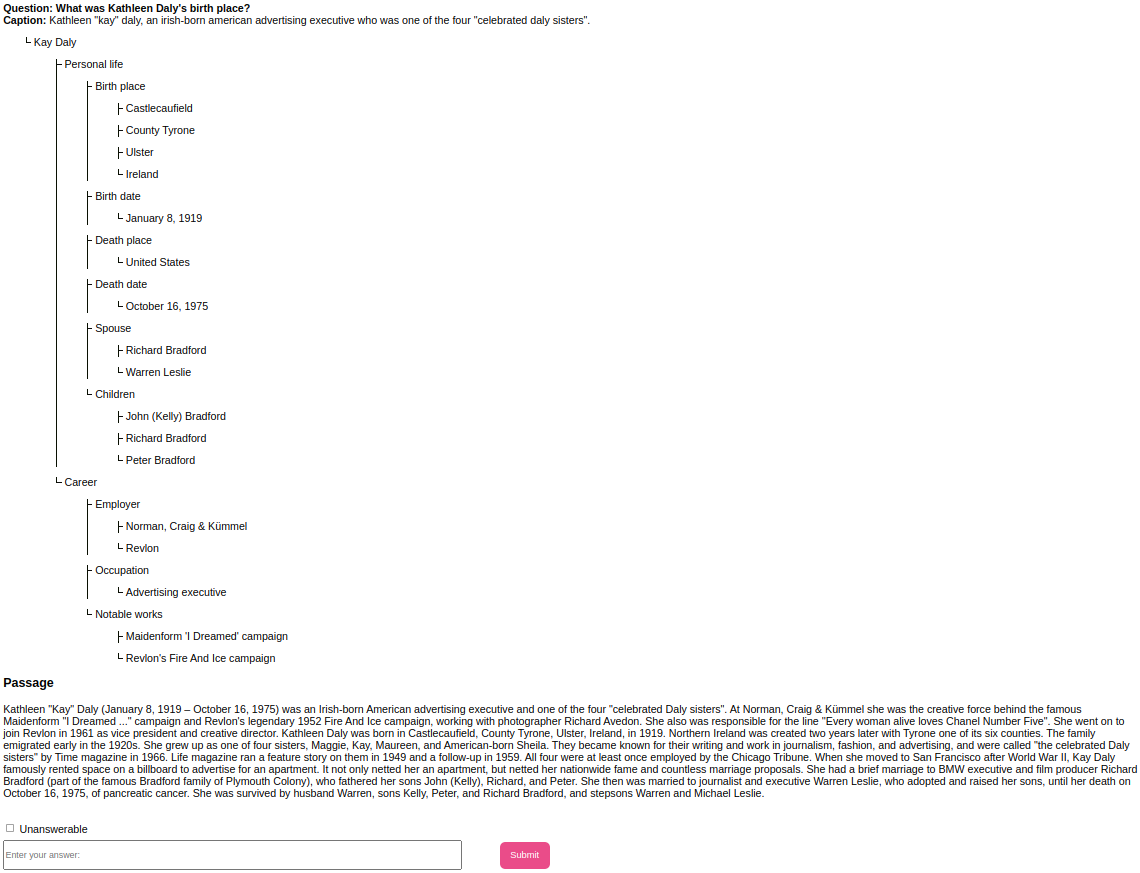}
    \caption{A followup frame shown after Figure~\ref{fig:user_study_initq} with structure (mind map) output and input text as context.}
    \label{fig:user_study_mm_s_text}
\end{figure*}

Usually, mind maps are represented as a graph as shown in Figure~\ref{fig:full_mindmap_example}. However, for the text comprehension user study described in Section~\ref{sec:user_study}, to avoid bias due to color or orientation, we simplify the representation as a tree (Figure~\ref{fig:user_study_mmonly}). To establish the known query intent, annotators' are first shown with input question, e.g., Figure~\ref{fig:user_study_initq}. Next, on clicking \texttt{Show content} button, annotators are shown context in the form of either text (Figure~\ref{fig:user_study_mmtext}), structure (Figure~\ref{fig:user_study_mmonly}), or structure + text (Figure~\ref{fig:user_study_mm_s_text}). 
The question-answer pairs were generated automatically conditioned on input text (Section~\ref{sec:autoqa}). Annotators were also allowed to mark an instance un-answerable. The user study for tables is performed in a similar manner.
We annotated $100$ question-answer pairs for both mind maps and tables. Each input instance is annotated with three different context combinations, leading to $600$ total annotations. We filtered instances that were marked un-answerable by the annotators (32\% and 22\% for tables and mind map study resp.). To avoid penalizing for spelling errors or other typing mistakes, the answers were evaluated via human evaluation. 

\par We adopt timed-comprehension for answering free-form questions as a proxy measure for usefulness of generated structured representation. This makes it different from categorical data annotations. We calculate rater agreement for the questionnaire responses. We find that 89.9\% of questions had full rater agreement regarding the correct response, with the only differences in the time taken to respond. Each question was shown to three raters. Table~\ref{tab:user_study_acc} shows the overall accuracy as percentage of questions answered correctly in different context. Irrespective of context combinations, annotators were able to answer the questions correctly with a high accuracy. 

\begin{table}[ht!]
    \centering
    \footnotesize
    \begin{tabular}{lc}
    \toprule
    \multicolumn{2}{c}{\textbf{Tables}}\\
    \midrule
    Table   & 95.6 \\
    Text    & 94.1 \\
    Table+Text & 94.1\\
    \bottomrule
    \end{tabular}
    \quad
    \begin{tabular}{lc}
    \toprule
    \multicolumn{2}{c}{\textbf{Mindmaps}}\\
    \midrule
     Mind map   & 97.7 \\
    Text    & 94.3 \\
    Mind map+Text &97.7\\
    \bottomrule
    \end{tabular}
    \caption{Answer accuracy (as percentage) for different context combinations. Structure context performs on par/better compared to text.}
    \label{tab:user_study_acc}
\end{table}

\section{Prompts}
\label{sec:prompts}
We include the different prompts used in this study. In our implementation we use Jinja~(\url{https://jinja.palletsprojects.com/}) to specify the prompt template. The prompts can be found in \Cref{fig:prompts/autoqa,fig:prompts/critic_factuality,fig:prompts/critic_global_structure,fig:prompts/critic_local_structure,fig:prompts/mindmap_reuse,fig:prompts/mindmap_simple_cot,fig:prompts/text_to_table_caption}.

\begin{figure*}
\begin{subfigure}{\textwidth}
   \texttt{The Mersey-class cruisers were improved versions of the Leander class with more armour and no sailing rig on a smaller displacement. Like their predecessors, they were intended to protect British shipping. The cruisers had a length between perpendiculars of 300 feet (91.4 m), a beam of 46 feet (14.0 m) and a draught of 20 feet 2 inches (6.1 m). They displaced 4,050 long tons (4,110 t). The ships were powered by a pair of two-cylinder horizontal, direct-acting, compound-expansion steam engines, each driving one shaft, which were designed to produce a total of 6,000 indicated horsepower (4,500 kW) and a maximum speed of 18 knots (33 km/h; 21 mph) using steam provided by a dozen cylindrical boilers with forced draught. The Mersey class carried enough coal to give them a range of 8,750 nautical miles (16,200 km; 10,070 mi) at a speed of 10 knots (19 km/h; 12 mph). The ships' complement was 300 to 350 officers and ratings. Their main armament consisted of two breech-loading (BL) 8-inch (203 mm) guns, one each fore and aft on pivot mounts. Their secondary armament was ten BL 6-inch (152 mm) guns, five on each broadside in sponsons. Protection against torpedo boats was provided by three quick-firing (QF) 6-pounder Hotchkiss guns and three QF 3-pounder Hotchkiss guns. The ship was also armed with a pair of submerged 14-inch (356 mm) torpedo tubes and carried a pair of 14-inch torpedo carriages. The Mersey-class ships were protected by a lower armoured deck that was 2 inches (51 mm) on the flat and 3 inches (76 mm) on the slope. It sloped down at the bow to reinforce the ram. The armoured sides of the conning tower were 9 inches (229 mm) thick.}
    \caption{Input text for table genetation.}
    \label{fig:table_input_text}
    \end{subfigure}
   
   \centering
   \begin{subfigure}{\textwidth}
    \includegraphics[width=\textwidth]{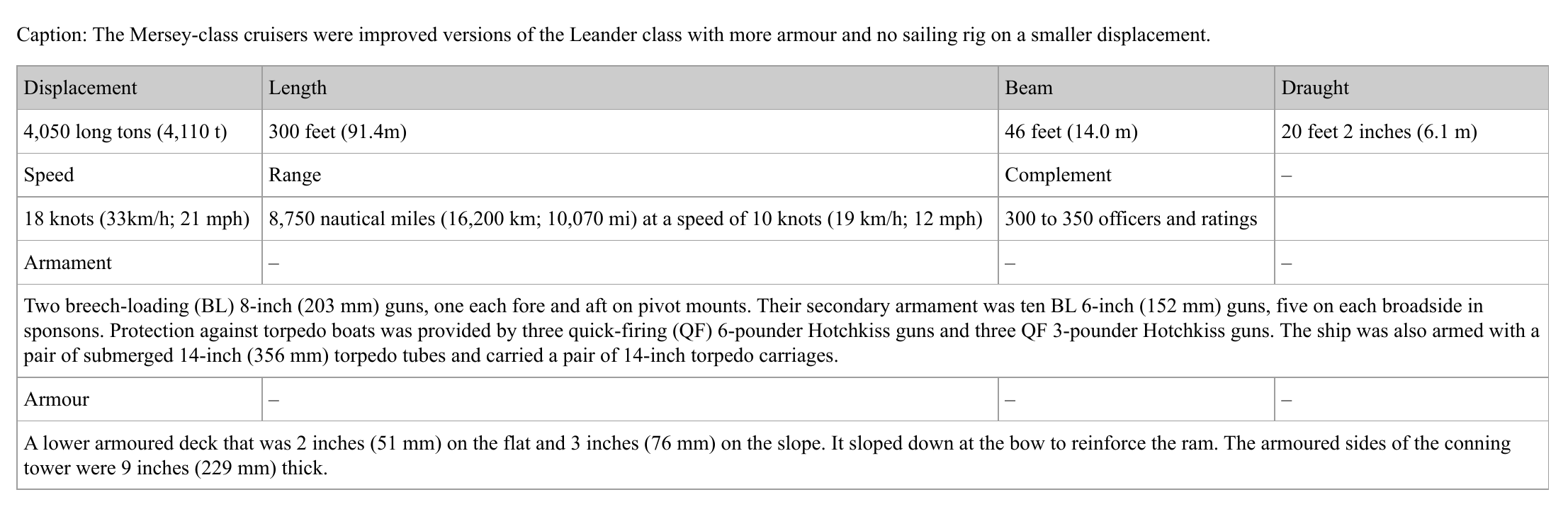}
    \caption{Single table generation output.}
    \label{fig:single_table_full}
    \end{subfigure}
    \begin{subfigure}{\textwidth}
    \includegraphics[width=\textwidth]{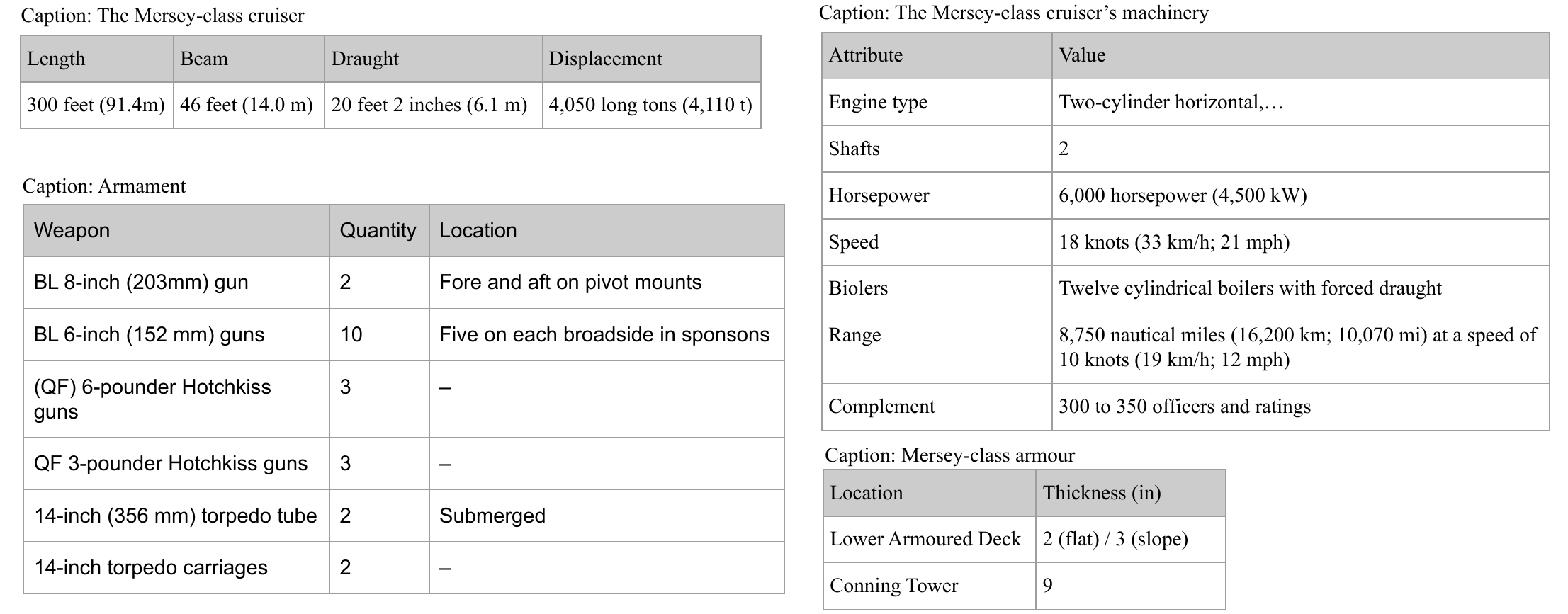}
    \caption{Multiple table generation output.}
    \label{fig:multi_table_full}
    \end{subfigure}
    
\caption{Example outputs for single and multiple table generation approach. Text in (a) shows the input. (b) and (c) show the outputs for single and multiple table generation respectively.}
\label{fig:full_table_example}
\end{figure*}


\begin{figure}
    \centering
\begin{lstlisting}[frame=single,basicstyle=\linespread{0.8}\ttfamily\tiny,breaklines=true,inputencoding=utf8,postbreak=\mbox{\textcolor{red}{$\hookrightarrow$}\space}]
{% set step1 = '{"node": "Global Climate change"}' -%}
{% set step2 = '{"node": "Global Climate change","branches":[{"node": "effects"},{"node": "causes"},{"node": "solutions"}]}' -%}
{% set step3 = '{"node": "Global Climate change","branches":[{"node": "effects","branches": [{"node": "melting ice"},{"node": "heat waves"}]},{"node": "causes","branches": [{"node": "Enhanced greenhouse effect"},{"node": "Pollution"}]},{"node": "solutions","branches": [{"node": "Individual efforts"},{"node": "International resolutions"}]}]}' -%}
{% set step4 = '{"node": "Global Climate change","branches":[{"node": "effects","branches": [{"node": "melting ice"},{"node": "heat waves","branches": [{"node" : "droughts"}]}]},{"node": "causes","branches": [{"node": "Enhanced greenhouse effect"},{"node": "Pollution","branches": [{"node": "Carbon emission"},{"node": "Burning coal"}]}]},{"node": "solutions","branches": [{"node": "Individual efforts"},{"node": "International resolutions"}]}]}' -%}
A mind map is a diagram used to visually organize information into a hierarchy, showing relationships among pieces of the whole. It is often created around a single concept.  Major ideas are connected directly to the central concept, and other ideas branch out from those major ideas. Mind maps can be generated based on the content present in text in multiple steps.
Consider the following example.
Given the following text:
Global climate change has many effects, including melting ice, heat waves, and droughts. It is caused by the enhanced greenhouse effect, which is caused by pollution, such as carbon emissions and burning coal. Solutions to global climate change include individual efforts and international resolutions.

Choose primary concept that is the root
Output:
MindMap
{{ format_json(step1)|safe }}
END_THOUGHT
Can we add branches ?
Output: Yes
END_THOUGHT
Add branches:
MindMap
{{ format_json(step2)|safe }}
END_THOUGHT
Can we add branches ?
Output: Yes
END_THOUGHT
Add branches:
MindMap
{{ format_json(step3)|safe }}
END_THOUGHT
Can we add branches ?
Output: Yes
END_THOUGHT
Add branches:
MindMap
{{ format_json(step4)|safe }}
END_THOUGHT
Can we add branches ?
Output: No
END_THOUGHT

Now for the text below:
{{input_text}}
Choose primary concept that is the root
Output:
MindMap{%- if root %}
{{root}}
END_THOUGHT
{% if current_mindmap -%}
Can we add branches?
Output: Yes
Add branches:
MindMap
{{current_mindmap}}
END_THOUGHT
{%- endif %}
Can we add branches ?
{%- if y_n_current %}
Output: {{y_n_current}}
END_THOUGHT
Add branches:
MindMap
{%- else %}
Output: {%- endif %}
{%- endif %}
\end{lstlisting}
    \caption{Iterative prompt in Jinja template format for mind map generation that is used in Algorithm~\ref{alg:mm}.}
    \label{fig:prompts/mindmap_reuse}
\end{figure}

\begin{figure}
    \centering
\begin{lstlisting}[frame=single,basicstyle=\linespread{0.8}\ttfamily\tiny,breaklines=true,inputencoding=utf8,postbreak=\mbox{\textcolor{red}{$\hookrightarrow$}\space}]
{% set step = '{"node": "Global Climate change","branches":[{"node": "effects","branches": [{"node": "melting ice"},{"node": "heat waves","branches": [{"node" : "droughts"}]}]},{"node": "causes","branches": [{"node": "Enhanced greenhouse effect"},{"node": "Pollution","branches": [{"node": "Carbon emission"},{"node": "Burning coal"}]}]},{"node": "solutions","branches": [{"node": "Individual efforts"},{"node": "International resolutions"}]}]}' -%}
A mind map is a diagram used to visually organize information into a hierarchy, showing relationships among pieces of the whole. It is often created around a single concept. Major ideas are connected directly to the central concept, and other ideas branch out from those major ideas. Mind maps can be generated based on the content present in text in multiple steps.
Consider the following example.
Given the following text:
Global climate change has many effects, including melting ice, heat waves, and droughts. It is caused by the enhanced greenhouse effect, which is caused by pollution, such as carbon emissions and burning coal. Solutions to global climate change include individual efforts and international resolutions.

Thought: Primary concept is Global climate change. Global climate change has branches, effects, causes and solutions. Effects have branches that include effects, melting ice and heat waves. Causes have branches enhanced greenhouse effect. Solutions have branches, individual efforts and international resolutions.

Output:
MindMap
{{ format_json(step)|safe }}

Now summarize the following text as a mind map.
{{input_text}}
\end{lstlisting}
    \caption{Prompt in Jinja template format for mind map generation without iterative process.}
    \label{fig:prompts/mindmap_simple_cot}
\end{figure}

\begin{figure}
    \centering
\begin{lstlisting}[frame=single,basicstyle=\linespread{0.8}\ttfamily\tiny,breaklines=true,inputencoding=utf8,postbreak=\mbox{\textcolor{red}{$\hookrightarrow$}\space}]
Your task is to divide a passage into smaller passages grouped by similar facts. Separate passages by __NEW_PASSAGE__.
For example giving the following passage:
On December 31, 2016, the bank met all capital adequacy requirements to which it was subject and exceeded the regulatory minimum capital levels to be considered well-capitalized under the regulatory framework for prompt corrective action. At December 31, 2016, the bank's ratio of common equity tier 1 capital to risk-weighted assets was 11.59%, total capital to risk-weighted assets was 12.85%, tier 1 capital to risk weighted assets was 11.59% and tier 1 capital to average assets was 10.10%. Our shareholders are entitled to dividends when and if declared by our board of directors out of funds legally available. Connecticut law prohibits us from paying cash dividends except from our net profits, which are defined by state statutes. On January 27, 2016 the company's board of directors declared a $0.05 per share cash dividend, payable February 22, 2016 to shareholders of record on February 12, 2016. On April 27, 2016 the company's board of directors declared a $0.05 per share cash dividend, payable May 26, 2016 to shareholders of record on May 16, 2016. On July 27, 2016 the company's board of directors declared a $0.05 per share cash dividend, payable August 26, 2016 to shareholders of record on August 16, 2016. The company's board of directors declared a $0.07 per share cash dividend, payable November 28, 2016 to shareholders of record on November 18, 2016, representing a 40% increase when compared to the last quarter.
Smaller passages looks like:
__NEW_PASSAGE__
On December 31, 2016, the bank met all capital adequacy requirements to which it was subject and exceeded the regulatory minimum capital levels to be considered well-capitalized under the regulatory framework for prompt corrective action. At December 31, 2016, the bank's ratio of common equity tier 1 capital to risk-weighted assets was 11.59%, total capital to risk-weighted assets was 12.85%, tier 1 capital to risk weighted assets was 11.59% and tier 1 capital to average assets was 10.10%.
__NEW_PASSAGE__
Our shareholders are entitled to dividends when and if declared by our board of directors out of funds legally available. Connecticut law prohibits us from paying cash dividends except from our net profits, which are defined by state statutes. On January 27, 2016 the company's board of directors declared a $0.05 per share cash dividend, payable February 22, 2016 to shareholders of record on February 12, 2016. On April 27, 2016 the company's board of directors declared a $0.05 per share cash dividend, payable May 26, 2016 to shareholders of record on May 16, 2016. On July 27, 2016 the company's board of directors declared a $0.05 per share cash dividend, payable August 26, 2016 to shareholders of record on August 16, 2016. The company's board of directors declared a $0.07 per share cash dividend, payable November 28, 2016 to shareholders of record on November 18, 2016, representing a 40% increase when compared to the last quarter.

Now divide the following passage into Smaller passages grouped by similar facts.
{{input_text}}
\end{lstlisting}
\begin{lstlisting}[frame=single,basicstyle=\linespread{0.8}\ttfamily\tiny,breaklines=true,inputencoding=utf8,postbreak=\mbox{\textcolor{red}{$\hookrightarrow$}\space}]
Summarize the contents of the text below in a table.

{{input_text}}

Use the following format.

Caption: A caption for the table you generate. Can be multiple lines
Table: A table in markdown format.

Caption:
\end{lstlisting}
    \caption{Text segmentation prompt (top) for multiple table generation. Zero-shot prompt for text to table and caption generation (bottom).}
    \label{fig:prompts/text_to_table_caption}
\end{figure}

\begin{figure}
    \centering
\begin{lstlisting}[frame=single,basicstyle=\linespread{0.8}\ttfamily\tiny,breaklines=true,inputencoding=utf8,postbreak=\mbox{\textcolor{red}{$\hookrightarrow$}\space}]
Given the text:
{{bullet_points}}

Table:
{{table}}

Rewrite the table adding a citation column, using the format [X], indicating the sentence number where that specific information can be found. When unsure use [NA].

Table with citations:
\end{lstlisting}
    
\begin{lstlisting}[frame=single,basicstyle=\linespread{0.8}\ttfamily\tiny,breaklines=true,inputencoding=utf8,postbreak=\mbox{\textcolor{red}{$\hookrightarrow$}\space}]
Input text:
{{bullet_points}}

Paths:
{{paths}}

Add an attribution to each path, using the format [X], where X is a sentence of the input text. When unsure use [NA] as attribution.

Paths with attribution:
\end{lstlisting}

    \caption{Factuality critic prompts for Table (top) and Mind maps (bottom).}
    \label{fig:prompts/critic_factuality}
\end{figure}

\begin{figure}
    \centering
\begin{lstlisting}[frame=single,basicstyle=\linespread{0.8}\ttfamily\tiny,breaklines=true,inputencoding=utf8,postbreak=\mbox{\textcolor{red}{$\hookrightarrow$}\space}]
Your task is to check if all the values in a list falls under a category. Go over all the values one by one and check if they belong to the assigned category.  Use the following format to answer.

Thought: Reasoning for the answer.
Answer: Single final answer yes or no.

Category: {{category}}
Values:
{{values}}

Thought:
\end{lstlisting}
    
\begin{lstlisting}[frame=single,basicstyle=\linespread{0.8}\ttfamily\tiny,breaklines=true,inputencoding=utf8,postbreak=\mbox{\textcolor{red}{$\hookrightarrow$}\space}]
There are some words or sentences that describes concept while other describes values associated with them. Values are defined as ordinals, type of job, degree, education level, location, region, date etc.
Answer No If any words is not a specific value otherwise answer yes.
For example:
Words:
Delhi
10
Cat
26 May
Lawyer

Thought: All the words are specific content words.
Answer: yes

Words:
IBM
Trucks
Birth
Family
26 May

Thought: Many words such as Trucks, Family, Birth are concept without specific values.
Answer: no


Words:
{{words}}

Thought:
\end{lstlisting}
    
    \caption{Local structure critic prompt for Tables (top, zero-shot) and Mind maps (bottom, few-shot).}
    \label{fig:prompts/critic_local_structure}
\end{figure}

\begin{figure}
    \centering
\begin{lstlisting}[frame=single,basicstyle=\linespread{0.8}\ttfamily\tiny,breaklines=true,inputencoding=utf8,postbreak=\mbox{\textcolor{red}{$\hookrightarrow$}\space}]
A TOC contains titles that are either too specific/long or too generic/short, or common sense. A useful TOC contains short, concise and informative titles at the right level of abstraction. Following a path in the TOC should allow you to generate or infer a sensible sentence.

Table of contents: root((K-26))
    1. southern terminus
      1.1. location
        1.1.2. city
        1.1.3. state
      1.2. state
    2. northern terminus
      2.1. location
        2.1.1. city
        2.1.2. state
      2.2. state
    3. maintained by
      3.1. organization
    4. traffic
      4.1. annual average daily traffic
      4.2. trucks
    5. national highway system
      5.1. listed
Thought (be specific): I can't create a sensible sentence following the path K-26 -> southern terminus -> location -> city
Useful: no
Table of contents: root((Assyria))
    1. Assyrian cities
      1.1. Aššur
      1.2. Nineveh
    2. Assyrian empire
      2.1. Neo-Assyrian Empire
    3. Assyrian period
    4. Assyrian kingdoms
      4.1. Adiabene
      4.2. Osroene
      4.3. Assur
      4.4. Beth Garmai
    5. Assyrian language
      5.1. Old Aramaic language
      5.2. Syriac language
Thought (be specific): All the titles contain useful information. All the paths allow generation of sensible sentences.
Useful: yes
Table of contents: root((Lonnie Johnson))
    1. Early life
      1.1. Birth
      1.2. Family
      1.3. Education
    2. Career
      2.1. Blues contest
      2.2. Recording contract
      2.3. Recordings
      2.4. Tours
      2.5. Collaborations
      2.6. Style
      2.7. Compositions
      2.8. Great Depression
      2.9. Later years
    3. Death
      3.1. Date
      3.2. Place
      3.3. Cause
Thought (be specific): Lonnie Johnson -> Early life -> Birth is a generic path not useful for generating a sentence.
Useful: no
Table of contents: {{toc}}
Thought (be specific):
\end{lstlisting}
    
    \caption{Global structure critic few shot prompt for Mind map.}
    \label{fig:prompts/critic_global_structure}
\end{figure}

\begin{figure}
    \centering
\begin{lstlisting}[frame=single,basicstyle=\linespread{0.8}\ttfamily\tiny,breaklines=true,inputencoding=utf8,postbreak=\mbox{\textcolor{red}{$\hookrightarrow$}\space}]
Your task is to generate a list of fact based questions that can be answered by the text passage. The format should be [Question][Answer].
Paragraph: {{text}}
\end{lstlisting}
\begin{lstlisting}[frame=single,basicstyle=\linespread{0.8}\ttfamily\tiny,breaklines=true,inputencoding=utf8,postbreak=\mbox{\textcolor{red}{$\hookrightarrow$}\space}]
Check if the following two answers are equivalent.
Use the following format.
Question: question text
Answer 1: answer text
Answer 2: answer text
Conclusion: Yes/No

Question: {{context_question}}
Answer 1: {{answer1}}
Answer 2: {{answer2}}
Conclusion:
\end{lstlisting}
\begin{lstlisting}[frame=single,basicstyle=\linespread{0.8}\ttfamily\tiny,breaklines=true,inputencoding=utf8,postbreak=\mbox{\textcolor{red}{$\hookrightarrow$}\space}]
Answer in concise manner the question using the information below. Say <unknown> when the questions cannot be answered.

{{data}}

Question:
{{question}}
\end{lstlisting}
    \caption{Prompts used by the AutoQA pipeline: QA pair generation prompt \textit{(top)}; Conditional answer equivalence \textit{(middle)}; Question answering prompt \textit{(bottom)}}
    \label{fig:prompts/autoqa}
\end{figure}

\end{document}